\documentclass[10pt,twocolumn,letterpaper]{article}
\usepackage{cvpr}
\usepackage{times}
\usepackage{epsfig}
\usepackage{graphicx}
\usepackage{amsmath}
\usepackage{amssymb}
\usepackage{subcaption}
\usepackage{balance}
\usepackage[dvipsnames]{xcolor}
\usepackage[utf8]{inputenc} %
\usepackage[T1]{fontenc}    %
\usepackage{url}
\usepackage{amsfonts}
\usepackage{nicefrac}       %
\usepackage{microtype}      %
\usepackage{lipsum}
\usepackage{epstopdf}
\usepackage{color}
\usepackage{verbatim}
\usepackage{pdfpages}
\usepackage{caption} 
\usepackage{float}
\usepackage{booktabs}
\usepackage{colortbl}
\usepackage{makecell}
\usepackage{mathtools}
\usepackage{tabularx}

\def\cvprPaperID{29} %

\cvprfinalcopy %
\ifcvprfinal\fi

\begin{document}

\title{Systematic Evaluation of Backdoor Data Poisoning Attacks on Image Classifiers}

\author{
Loc Truong$^{1}$, Chace Jones$^{1}$, Brian Hutchinson$^{1,2}$, Andrew August$^{2}$, \\ Brenda Praggastis$^{2}$, Robert Jasper$^{2}$, Nicole Nichols$^{2}
$, Aaron Tuor$^{2}$\\
$^{1}$Western Washington University,
$^{2}$Pacific Northwest National Laboratory\\
{\tt\small \{truongl, jonesc48, hutchib2\}@wwu.edu, firstname.lastname@pnnl.gov}}

\maketitle

\begin{abstract}
Backdoor data poisoning attacks have recently been demonstrated in computer vision research as a potential safety risk for machine learning (ML) systems. Traditional data poisoning attacks manipulate training data to induce unreliability of an ML model, whereas backdoor data poisoning attacks maintain system performance unless the ML model is presented with an input containing an embedded ``trigger'' that provides a predetermined response advantageous to the adversary.
Our work builds upon prior backdoor data-poisoning research for ML image classifiers and systematically assesses different experimental conditions including types of trigger patterns, persistence of trigger patterns during retraining, poisoning strategies, architectures (ResNet-50, NasNet, NasNet-Mobile), datasets (Flowers, CIFAR-10), and potential defensive regularization techniques (Contrastive Loss, Logit Squeezing, Manifold Mixup, Soft-Nearest-Neighbors Loss). 
Experiments yield four key findings. 
    First, the success rate of backdoor poisoning attacks varies widely, depending on several factors, including model architecture, trigger pattern and regularization technique.
    Second, we find that poisoned models are hard to detect through performance inspection alone. 
    Third, regularization typically reduces backdoor success rate, although it can have no effect or even slightly increase it, depending on the form of regularization. 
    Finally, backdoors inserted through data poisoning can be rendered ineffective after just a few epochs of additional training on a small set of clean data without affecting the model's performance.
\end{abstract}

\section{Introduction}
As deep learning models become more ubiquitous we must assess the safety of the machine learning model development process. Machine learning attack scenarios can be broadly split into two types \cite{UCBAdversarial}.
In a {\bf {\em causative attack}} an adversary embeds flaws into model behavior by design during model development.
In contrast, in an {\bf {\em exploratory attack}} an adversary develops or discovers inputs on which the model will make unexpected errors. 
Exploratory attack scenarios dominate the research publications \cite{carlini2017towards, szegedy2013intriguing, biggio2018wild}, while backdoor data poisoning is a recently introduced causative attack that can allow adversaries to induce specific model errors. 
Backdoor data poisoning is an adversarial manipulation of training data and labels, to create a {\bf {\em backdoor}} which allows the model to respond to a {\bf {\em trigger-pattern}}, but otherwise operate normally.
Backdoor poisoning can be introduced by modifying not only the training data \cite{DBLP:journals/corr/abs-1708-06733}, but also the training procedure \cite{DBLP:journals/corr/abs-1807-00459}, or by direct manipulation of the model weights or architecture \cite{DBLP:journals/corr/abs-1812-03128}.
This work assesses computer vision classifiers across a range of modeling choices and backdoor data poisoning strategies that manipulate training images and labels, and provides suggestions for defense and mitigation. 

\paragraph{Threat Model}

Deep learning models are being used to solve a wide range of problems including image recognition \cite{ILSVRC15,DBLP:journals/corr/ZophVSL17}, machine translation \cite{bahdanau2014neural,DBLP:journals/corr/WuSCLNMKCGMKSJL16}, and speech recognition \cite{DBLP:journals/corr/abs-1303-5778,DBLP:journals/corr/abs-1712-01769}. 
The current prevailing trend in deep learning development cycles is to pre-train models from a large public dataset and then fine-tune on a smaller internal proprietary dataset. 
Deployed systems using classification models built from public data with uncertain provenance may pose safety risks due to potential data poisoning \cite{DBLP:journals/corr/abs-1708-06733}. 

In this research, we use the scenario of a potentially poisoned public dataset to evaluate model development choices. 
This scenario is designed around a trigger-pattern in a subset of images in the public dataset. 
These training images, embedded with trigger patterns, are re-labeled to the adversary's chosen prediction label. 
A successful attack occurs when a deployed model, trained on the poisoned dataset, behaves normally when encountering natural images but produces the adversary's chosen label when presented images with embedded triggers. 

\paragraph{Contributions}
Backdoor methods have been demonstrated on numerous datasets and model architectures. 
Typical domains include face recognition \cite{chen2017targeted, liu2017trojaning, ji2018model}, self-driving cars \cite{DBLP:journals/corr/abs-1708-06733, liao2018backdoor, DBLP:journals/corr/abs-1902-11237, liu2017trojaning}, medical applications \cite{ji2018model}, and standard benchmarks \cite{DBLP:journals/corr/abs-1708-06733, turner2018clean, shafahi2018poison, DBLP:journals/corr/abs-1807-00459, DBLP:journals/corr/abs-1806-07409}.
Because a common poisoning methodology has not been established, it is not possible to directly compare results for attack demonstrations across  datasets and architectures from different research publications. 
Our current work addresses this limitation by performing experiments across a broad matrix of conditions.
We systematically evaluate key factors which may affect the success and persistence of the backdoor attack.
These key factors include the model architecture, the adversary's trigger pattern, poisoning strategy, the dataset and associated classification task.
Our experimental results show these factors can greatly impact backdoor data poisoning attacks.

Defense and  mitigation of backdoor data poisoning is also assessed through both regularization during training and a series of experiments where small amounts of clean data are used to fine-tune a trained (poisoned) model. 
We demonstrate that, across a range of models, without specific knowledge of poisoning methods, 
a defender can significantly diminish backdoor attack effects by fine-tuning the model on a trusted source of known, clean data.  

\section{Experiment Matrix}
\begin{figure}
\centering
\includegraphics[width=0.42\textwidth, height=1.8in]{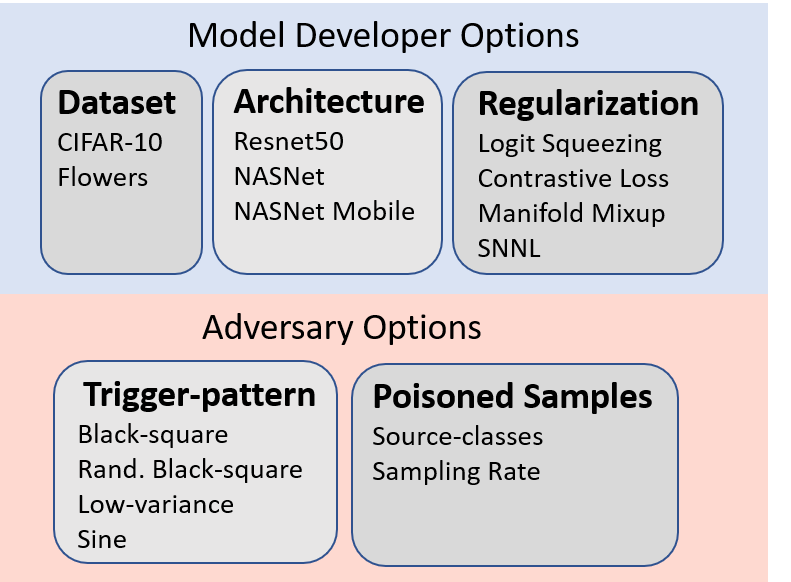}
\caption{Factors systematically varied in our experiments.}
\label{fig:factors}
\end{figure}

There are a wide range of factors and associated values that may affect the success of backdoor data poisoning attacks. Some factors are directly under the control of the model developer, whereas others are associated with the adversary's poisoning method. Figure \ref{fig:factors} shows the factors and range of associated values used in our experiments. This section describes each factor and associated values and motivates their selection for the present study.
\label{sec:ExperimentalMethods}

\subsection{Dataset}
\label{subsection:Datasets}

We assess backdoor data poisoning strategies on two datasets to compare possible effects of dataset selection on attack success. First is the Flowers dataset\footnote{https://www.kaggle.com/alxmamaev/flowers-recognition} containing 4,242 $224 \times 224$ pixel images from five different types of flowers. Second is the CIFAR-10 dataset \cite{krizhevsky2014cifar}
containing 50,000 $32 \times 32$ pixel thumbnail images across ten classes. By employing these two datasets we can compare results on CIFAR-10 to results on higher resolution images which are typical for image classification systems.
\subsection{Model Architectures}
For this study, we choose three state of the art computer vision classifiers that are widely used in deployed settings across application domains; namely, 
ResNet-50 \cite{DBLP:journals/corr/HeZRS15}, NasNet \cite{DBLP:journals/corr/ZophVSL17}, and NasNet-Mobile \cite{DBLP:journals/corr/ZophVSL17}. We initialize each model with publicly available ImageNet \cite{imagenet_cvpr09} weights. We then fine tune them with Flowers and CIFAR-10 data during training.

\subsection{Regularization Techniques}
The success of recent backdoor detection methods \cite{chen2018detecting, xiang2019benchmark, soremekun2020exposing} and exploratory attack defensive measures \cite{gorbett2020utilizing, miller2020adversarial} which analyze the latent space of deep learning models suggest that latent space regularization may have significant effect on backdoor attack success.
With image height and width ($H$, $W$), a generic classifier can be defined as a composition of functions $f = g \circ h: \mathbb{R}^{H \times W\times 3} \rightarrow \mathbb{R}^n$, mapping an image to a class distribution over $n$ classes. The intermediate function $h$ maps the image to the final hidden representation of the classifier, and $g$ is a multinomial logistic regression classifier that maps hidden representations to class probabilities. We also define $f_L({\bf x})$ as the logit output (non-normalized log probabilities) of the network prior to the final softmax activation. Our experiments compare backdoor attack performance on models trained using one of four regularization methods designed to constrain the latent space of the final hidden layer or classification logits of the image classifier.

\paragraph{Logit squeezing}
\cite{kannan2018adversarial} introduced 
logit-squeezing regularization as a method to provide model robustness to adversarial examples. For a training image, ${\bf x}$, Logit-squeezing adds $\mathcal{L}_{\texttt{LS}}= ||f_L({\bf x})||_2$ to the loss function to minimize the $l_2$ norm of the logit vector. 

\paragraph{Manifold Mixup}
Introduced in \cite{verma2018manifold}, Manifold mixup (MIXUP) attempts to fill in gaps in the latent space manifold by interpolating the latent representations and corresponding predictions.   
Pairs of image hidden representations from the minibatch $(h({\bf x}), h({\bf x}'))$ are averaged according to a randomly sampled mixing weight $\gamma \sim Uniform(0,1)$.  The loss function to train the classifer is then the cross-entropy, $\mathcal{H}$, between the network's prediction for interpolated hidden state pairs and the $\gamma$ weighted average of true one-hot class label distributions $({\bf y}, {\bf y}')$:

\begin{eqnarray}
\mathcal{L}_{\texttt{mix}} =&\mathcal{H}(g({\bf h}_{\texttt{mix}}), {\bf y}_{\texttt{mix}})\\
{\bf h}_{\texttt{mix}} =&(1 - \gamma) h({\bf x}') + \gamma h({\bf x})\\
{\bf y}_{\texttt{mix}} =&(1 - \gamma) {\bf y}' + \gamma {\bf y}
\end{eqnarray}

\paragraph{Contrastive Loss}
Contrastive loss \cite{chopra2005learning} encourages hidden representations from the same object class to be close together, and hidden representations from different object classes to be far apart. 
Let ${\bf x}$ and ${\bf x}'$ be two images. The contrastive regularization $L_{\texttt{contrast}}$ is: 
\begin{equation}
\frac{1}{n}\|h({\bf x})- h({\bf x}')\|_2 
\end{equation}
if  ${\bf x}$ and ${\bf x}'$ are the same class, and otherwise: 
\begin{equation}
\frac{n-1}{n}\max(0, c - \|h({\bf x})- h({\bf x}')\|_2) 
\end{equation}

\paragraph{Soft Nearest Neighbors Loss}
Soft Nearest Neighbors Loss (SNNL) \cite{frosst2019analyzing} regularization was introduced to improve hidden space representations in many settings.
SNNL weights the contribution of a pair of samples in a batch relative to the probability of being picked randomly as a nearest neighbor.
With batch samples $({\bf x}^{(i)}, {\bf y}^{(i)}), i={1, ..., b}$. 
and temperature $T$, the SNNL regularization term is:
\begin{equation}
\mathcal{L}_{snn} = -\log \Bigg(\frac{\sum_{j\ne i, {\bf y}^{(i)} \ne {\bf y}^{(j)}} e^{-\frac{\|h({\bf x}^{(i)})-h({\bf x}^{(j)})\|^2}{T}}}{\sum_{k \ne i} e^{-\frac{\|h({\bf x}^{(i)})-h({\bf x}^{(k)})\|^2}{T}}}\Bigg)
\end{equation}

\subsection{Trigger Patterns}
 In this work, the backdoor is embedded in a model via data poisoning with trigger patterns embedded in adversarially re-labeled images. 
Let ${\textbf x}  \in \mathbb{R}^{H \times W \times 3}$ be a training set image,  let $\alpha \in [0, 1]$ be the transparency of the trigger, and let ${\textbf m}  \in \{0, 1\}^{H \times W \times 3}$ be a mask with 1's in pixel positions the trigger will not alter. We introduce a trigger function $\mathcal{T}$ which returns a trigger ${\textbf t}$. $\mathcal{T}$ may be constant, draw a random sample from a distribution of triggers (\eg, augmentation or perturbation of a trigger template), or depend on ${\textbf x}$  in the case of an adaptive trigger. The general form for constructing a poisoned sample image, {\textbf p}, with an embedded trigger is then:
\begin{equation}
 {\textbf p} = ((1 - \alpha ) {\textbf x} + \alpha \mathcal{T}({\textbf x}))\odot ({\bf 1} - {\textbf m}) + {\textbf x}\odot {\textbf m}
\end{equation}
where $\odot$ is the elementwise multiplication and boldface {\bf 1} is an all-ones tensor of the same dimension as the image. 

Four trigger types are experimentally evaluated, low-variance (LV), sine-wave (SIN), black square (BS), and random square (RS). Within a single experiment scenario, the same trigger type is applied to all poisoned samples.
The black square trigger pattern is a 22 pixel square, located 22 pixels from both the top and left sides of the image. This is similar to the triangle checkerboard trigger used in \cite{DBLP:journals/corr/abs-1708-06733}. The random square trigger is the same as the black square but placed at a random rather than fixed location in the image. 
The low-variance trigger pattern introduced in \cite{DBLP:journals/corr/abs-1806-07409} is constructed with reference to a particular dataset to be poisoned. First a PCA decomposition is performed on the training data. Then an image not present in the training data is projected onto the last principal components that explains $\geq 0.5$ percent of the variance in the dataset. This projection is then mapped back into the original image space to form the trigger pattern. %
The sine trigger, introduced in \cite{DBLP:journals/corr/abs-1902-11237}, consists of gray scale pixel intensities which vary horizontally across the image according to a sine function.
In particular the value for all three channels at pixel $(i, j)$ for the sine trigger is 
   $ 0.4\sin(0.05\pi j).$

\begin{figure*}
\centering
\begin{tabular}{cccccc}
\subcaptionbox{Square $\alpha = 1$}{\includegraphics[width = 0.8in]{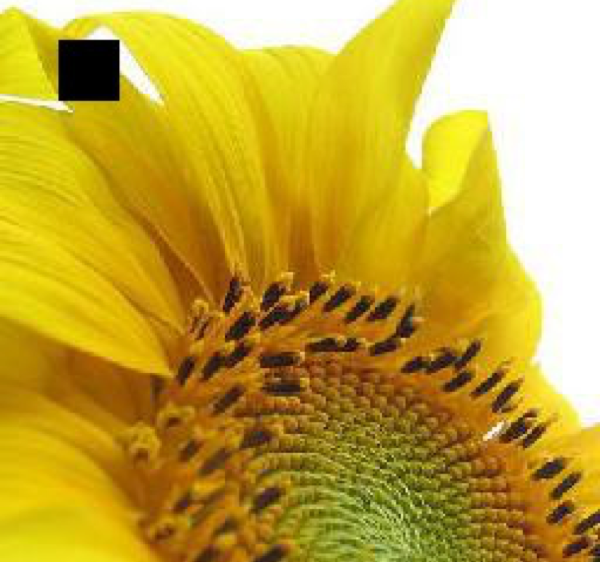}} &
\subcaptionbox{Random Square $\alpha = 1$}{\includegraphics[width = 0.8in]{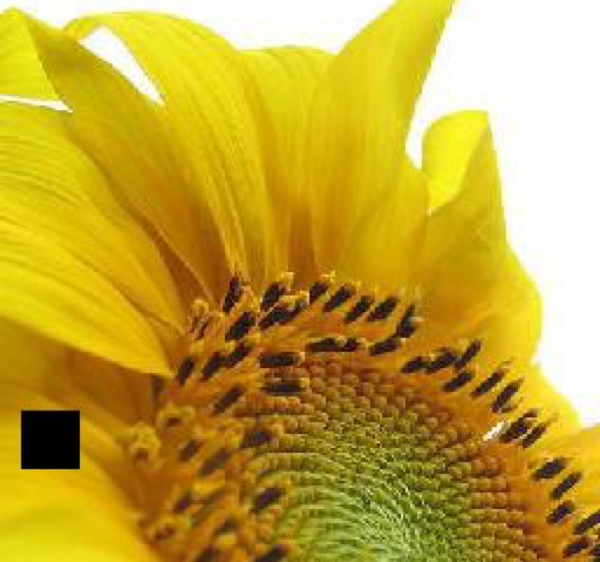}} &
\subcaptionbox{Sine $\alpha = 0.1$}{\includegraphics[width = 0.8in]{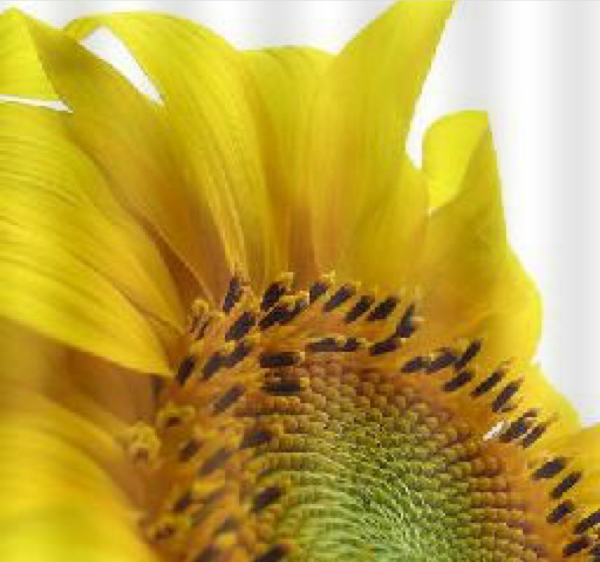}} &
\subcaptionbox{Sine $\alpha = 0.5$}{\includegraphics[width = 0.8in,height=0.75in]{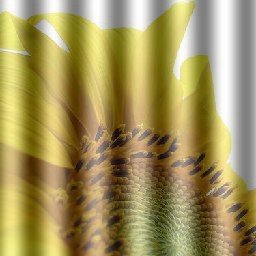}} &
\subcaptionbox{Low Variance $\alpha = 0.1$}{\includegraphics[width = 0.8in,height=0.75in]{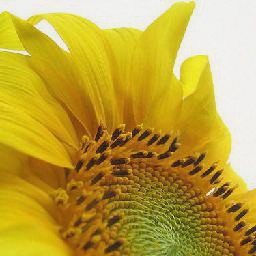}} &
\subcaptionbox{Low Variance $\alpha = 0.5$}{\includegraphics[width = 0.8in]{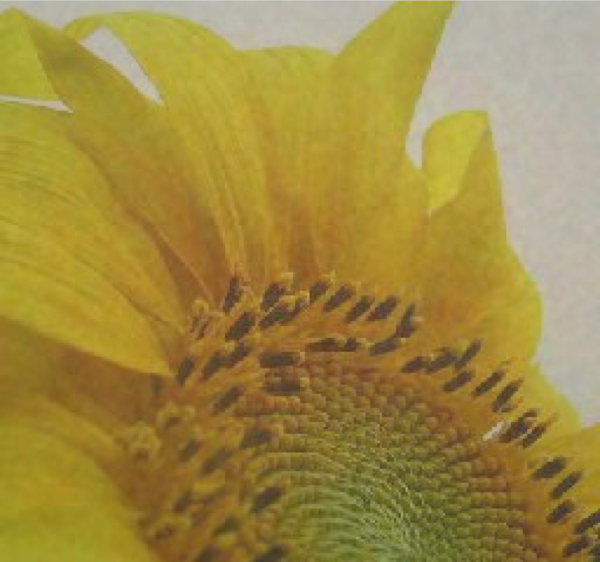}} 
\end{tabular}
\caption{Trigger patterns applied to an image from the Flowers dataset.}
\label{TriggerPatterns}
\end{figure*}

Trigger patterns that overlay the entire image such as sine and low variance in particular are easy to detect if their $\alpha$ values are too high. 
Considering this, we pay particular attention to a set of experimental runs with $\alpha$ values of 0.5 and 0.1 for the low variance and sine triggers respectively. These $\alpha$ values were selected as the highest alpha value before the image alteration becomes completely apparent. For the black square trigger we use an $\alpha$ value of 1 since it is relatively inconspicuous, covering a small portion of the image. Figure \ref{TriggerPatterns} shows an image from the flowers dataset with triggers embedded with these particular $\alpha$ values.  

\subsection{Poisoned Samples}
In addition to choice of trigger pattern, an adversary also has control over which images from the training dataset to poison (embed the trigger pattern). The {\bf {\em source-class}} is the true class of an image upon which a trigger is embedded, and the {\bf {\em poison-class}} is the class label given by the adversary. 
In the poisoning procedure we investigate, samples are drawn from the set of source-classes, embedded with a trigger pattern, and these poisoned samples then {\em supplant} clean samples from the poison-class. The untampered versions of the poisoned images remain in the source-classes. This method of poisoning ensures the number of images with each class label remains the same after poisoning, thereby eliminating class distribution shift due to data poisoning.  

An important factor which may affect the success of data poisoning is the distribution of poisoned images within the poisoned dataset.  
We define the {\bf {\em poison-rate}} as the percentage of the poison-class images replaced by poisoned samples.   
Let $N_1, N_2, ..., N_n$, be the number of images from each class in the training set, and $t$ be the index of the poison-class.
Given poison-rate $\lambda$, $\lfloor \lambda N_t \rfloor$ is the total number of samples to be replaced in the poison-class. For a set of source-classes $\mathcal{K} \subset \{1,..., n\}\setminus \{t\}$, the expected number of samples, $P_c$, drawn from each source-class, $c$, is:
\begin{equation}
    P_c = \lfloor \lambda N_t \rfloor \frac{N_c}{\sum_{k\in \mathcal{K}} N_k}
\end{equation}
The {\bf {\em effective-poisoning-rate}}, $p$, is defined as the percentage of the total number of training samples which are poisoned:
\begin{equation}
    p =   \frac{ \lfloor \lambda N_t \rfloor}{\sum_{k \in {\cal K}}^n N_k}
\end{equation}

The choice of source-classes has a direct effect on the distribution of poisoned images and so in addition to testing the effectiveness over various poison-rates we consider poisoning strategies which draw from a single source-class ({\bf {\em one-to-one}}) or multiple source classes ({\bf {\em many-to-one}}).
In one-to-one poisoning, poisoned images from a single source-class supplant images from a single poison-class.
In many-to-one poisoning, all classes excluding the poison-class are source-classes.
Table \ref{tab:sampling} shows class distribution and poison sample distribution statistics for the Flowers dataset with the many-to-one poisoning strategy and a poison-rate $\lambda =0.1$.

\begin{table}
\centering
\resizebox{0.46\textwidth}{!}{%
\begin{tabular}{lccccccc}
\rowcolor[HTML]{FFFFFF} 
 \toprule
 &    & {\bf $N_1$} & {\bf $N_2$} & {\bf $N_3$} &   {\bf $N_4$}&  {\bf $N_5$} &\\
  \midrule
  \rowcolor[HTML]{E5E5E5} 
 &    & 710 & 980 & 734 &   675&  904 &\\
 \toprule
{\bf $t$} &  {\bf $\lfloor \lambda N_t \rfloor$}  & {\bf $P_1$} & {\bf $P_2$} & {\bf $P_3$} &   {\bf $P_4$}&  {\bf $P_5$} & $p$\\
 \midrule
\rowcolor[HTML]{E5E5E5} 
1-daisy & 71 & 0 & 21.1 & 15.8  &14.6  &19.5 & 0.018\\
2-dandelion & 98 & 23.0 & 0 &23.8  & 21.9 & 29.3 & 0.025\\
\rowcolor[HTML]{E5E5E5} 
3-rose & 73 & 15.9 & 21.9 & 0 & 15.1 & 20.2 & 0.018\\
4-sunflower& 67 & 14.3 & 19.7 & 14.8 & 0 & 18.2 & 0.017\\
\rowcolor[HTML]{E5E5E5} 
5-tulip & 90 & 20.6 & 28.5 & 21.3 & 19.6 & 0 &0.023
\end{tabular}}
\caption{Poison class statistics with $\lambda = 0.1$ for many-to-one poisoning on the Flowers dataset.}
\label{tab:sampling}
\end{table}

\section{Experimental Setup}
\begin{figure}
\begin{center}
\includegraphics[width=0.47\textwidth]{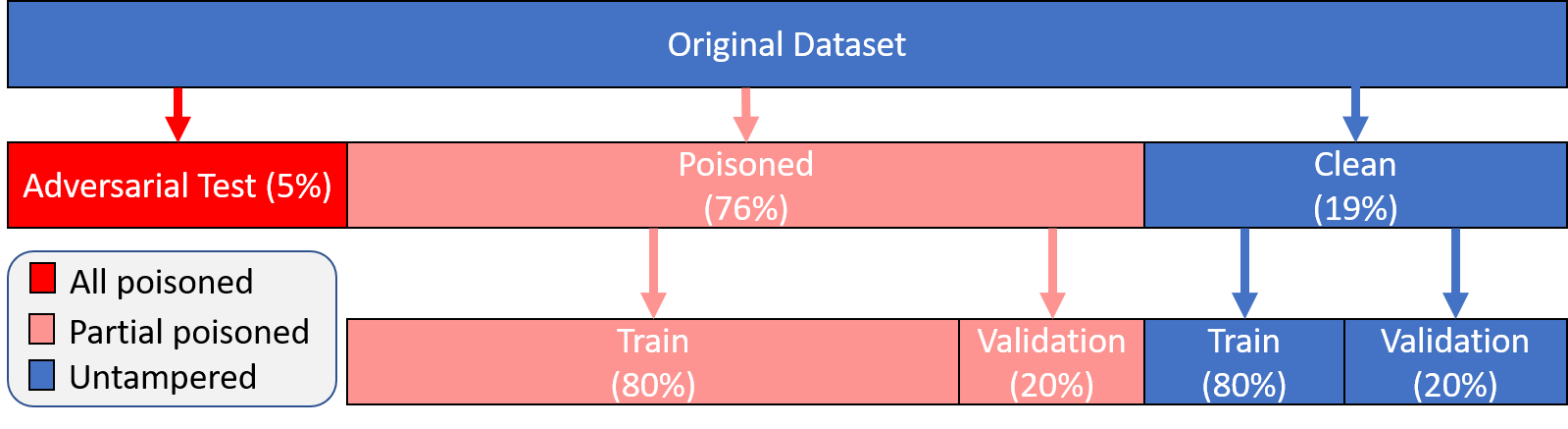}
\end{center}
\caption{Dataset partitions, where all (adversarial), some (poisoned) or no (clean) images have been poisoned.
}\label{fig:partition}
\end{figure}

\paragraph{Data Partitioning}
Because the goal of this research is to assess the overall safety of a model, we partition the data to allow performance evaluation from both adversary and model developer perspectives. Adversarial success rate (the fraction of poisoned images predicted to be the poison-class) is used to evaluate the adversary's success, while model accuracy is used to assess the model developer's. The dataset partitioning is shown in Figure \ref{fig:partition}. The original dataset is partitioned into a 76/19/5 split. In our experiment, the largest partition (76\%), which we call the {\bf {\em poison-set}}  plays the role of a larger, publicly available dataset that the adversary has tampered with, and that the model developer uses to train their first-pass computer vision model. The next largest partition (19\%), which we call the {\bf {\em clean-set}}, simulates a smaller internal dataset curated by the model developer to fine-tune the first-pass computer vision model. Note that the clean-set is $1/5$th the size of the poison-set. Both the clean-set and poison-set are further split into respective 80/20 train/validation sets. We use the remaining 5\% of the original dataset, which we call {\bf {\em adversarial-test}} to evaluate the success rate of the adversary. Accordingly, all images in the adversarial test set are poisoned. 
\paragraph{Poisoning details} Preliminary results showed higher adversarial success rate when poison-class samples were not corrupted, thus when constructing the poison-set, the trigger pattern is not embedded onto samples drawn from the poison-class (\ie, the poison-class is never one of the source-classes).  The adversarial test set also contains no images from the poison-class, since the purpose of the adversarial test set is to gauge the adversary's ability to {\em change} a prediction. To eliminate performance effects associated with changes in class distributions, we maintain the same number of samples from each class prior to and post poisoning. To ensure this consistent class size across all experimental runs, poisoned samples are exchanged for samples in the poison-class, but their non-poisoned counterparts are not removed from source-class which they are drawn from. 

\paragraph{State-of-the-art accuracy} Due to the data splits needed to conduct our study (Fig.~\ref{fig:partition}), our models only have access to around 60\% of the original data for training. As expected, these models do not achieve the state-of-the-art of models trained on the full training set. Ultimately, our goal is not state-of-the-art performance, but a systematic comparison of data poisoning; that said, we do tune each model to achieve as competitive of performance as possible. As a sanity check on the correctness of our training process, we successfully replicated publicly reported results for each of our models using the full training set. 

\paragraph{Training procedure}
Our procedure simulates the scenario where a model developer trains a base model on poisoned public data until the early stopping criterion (5 epochs with no improvement on validation accuracy) and then fine-tunes on an internal clean training set for a fixed number of epochs. During training, we monitor the model prediction accuracy on the clean and poison validation sets, and the adversarial success rate on the adversarial test set.
For each experimental run we perform independent random splits and poison samples at the specified rate randomly.

\section{Experimental Results and Analysis}
\label{sec:ResultsAnalysis}
In this section we analyze experimental results to answer several questions about backdoor attack success rate, backdoor persistence, and backdoor effects on model validation accuracy. Unless otherwise stated, the experiments described below use the ``many-to-one'' poisoning strategy, set poison-rate $\lambda=0.1$, trigger pattern transparency $\alpha=1$ for the Square and RS triggers patterns, $\alpha=0.1$ for sine and $\alpha=0.5$ for low variance.

\subsection{Effect of Trigger Pattern and Model}
We first analyze the effect of trigger patterns on different model architectures for backdoor poisoning. On the Flowers and CIFAR-10 datasets, we range over all trigger patterns, classes as poison-class, and architectures (180 runs total). 
We report average adversarial success rate and validation accuracy (over all classes) at early stopping after training on the poisoned training set. 
The average early stopping epoch for ResNet50, NasNet, and NasNet-Mobile was 14.6, 17.35, and 26.5, respectively.
The resulting adversarial successes are shown in Table~\ref{tab:main_table_advsucc} (see ``Retrained? No'' rows). It reveals that the square and random square triggers are the most effective for the Flowers dataset, while the sine and square triggers are the most effective for CIFAR-10. 
It also shows that NasNet-Mobile is by far the most robust to poisoning on Flowers, while NasNet-Mobile and NasNet are both slightly more robust on CIFAR-10.
Alarmingly, multiple combinations of model and trigger pattern yield adversarial success rates exceeding 60\%. 

Table~\ref{tab:main_table_accuracy} shows the model accuracy on the poisoned and clean validation sets (again see ``Retrained? No'' rows). For the models trained on Flowers, there is a negative correlation between model accuracy and robustness to poisoning, but for CIFAR-10 same models yield top performance on both. It is important to note that while the particular trigger pattern makes a significant difference in adversarial success, it has very little effect on the accuracy of the trained model, regardless of dataset.  Lastly, the minimal gap between performance on the poisoned and clean validation sets is an unfortunate finding for the model developer's perspective, because it suggests that poisoned data may be hard to detect by inspection of model performance.

\begin{table*}
\begin{center}
\resizebox{0.86\textwidth}{!}{%
\begin{tabular}{ccc|cccc|cccc|cccc|}
\cline{4-15}
 & & & \multicolumn{4}{c|}{\textbf{ResNet50}} & \multicolumn{4}{c|}{\textbf{NasNet}} & \multicolumn{4}{c|}{\textbf{NasNet Mobile}} \\ \cline{1-15} 
\multicolumn{1}{|c}{Dataset} & \multicolumn{1}{c}{Split} & \multicolumn{1}{c|}{Retrained?} & Square & RS & Sine & LV & Square & RS & Sine & LV & Square & RS & Sine & LV \\ \hline
\multicolumn{1}{|c}{Flowers} & Adversarial Test & No  & 0.75 & 0.64 & 0.24 & 0.26 & 0.65 & 0.58 & 0.18 & 0.06 & 0.33 & 0.15 & 0.14 & 0.12 \\ \hline
\multicolumn{1}{|c}{Flowers} & Adversarial Test & Yes & 0.08 & 0.09 & 0.06 & 0.05 & 0.18 & 0.14 & 0.06 & 0.04 & 0.05 & 0.05 & 0.06 & 0.06 \\ \hline
\multicolumn{1}{|c}{CIFAR-10} & Adversarial Test & No & 0.74 & 0.61 & 0.90 & 0.55 & 0.74 & 0.53 & 0.63 & 0.06 & 0.67 & 0.43 & 0.79 & 0.16 \\ \hline
\multicolumn{1}{|c}{CIFAR-10} & Adversarial Test & Yes & 0.04 & 0.04 & 0.06 & 0.05 & 0.09 & 0.08 & 0.08 & 0.02 & 0.05 & 0.03 & 0.08 & 0.05 \\ \hline
\end{tabular}}
\caption{Adversarial success before and after clean retraining, for Flowers and CIFAR-10.}
\label{tab:main_table_advsucc}
\end{center}
\end{table*}

\begin{table*}
\begin{center}
\resizebox{0.86\textwidth}{!}{%
\begin{tabular}{ccc|cccc|cccc|cccc|}
\cline{4-15}
 & & & \multicolumn{4}{c|}{\textbf{ResNet50}} & \multicolumn{4}{c|}{\textbf{NasNet}} & \multicolumn{4}{c|}{\textbf{NasNet Mobile}} \\ \cline{1-15} 
\multicolumn{1}{|c}{Dataset} & \multicolumn{1}{c}{Split} & \multicolumn{1}{c|}{Retrained?} & Square & RS & Sine & LV & Square & RS & Sine & LV & Square & RS & Sine & LV \\ \hline
\multicolumn{1}{|c}{Flowers} & Poisoned & No  & 0.89 & 0.87 & 0.85 & 0.87 & 0.87 & 0.87 & 0.85 & 0.85 & 0.81 & 0.80 & 0.79 & 0.80 \\ \hline
\multicolumn{1}{|c}{Flowers} & Clean & No & 0.88 & 0.87 & 0.87 & 0.89 & 0.87 & 0.87 & 0.87 & 0.87 & 0.83 & 0.83 & 0.81 & 0.83 \\ \hline
\multicolumn{1}{|c}{Flowers} & Poisoned & Yes & 0.86 & 0.85 & 0.86 & 0.86 & 0.86 & 0.87 & 0.87 & 0.86 & 0.80 & 0.79 & 0.80 & 0.80 \\ \hline
\multicolumn{1}{|c}{Flowers} & Clean & Yes & 0.89 & 0.89 & 0.89 & 0.90 & 0.87 & 0.88 & 0.89 & 0.89 & 0.81 & 0.82 & 0.82 & 0.84 \\ \hline
\multicolumn{1}{|c}{CIFAR-10} & Poisoned & No & 0.73 & 0.74 & 0.69 & 0.74 & 0.93 & 0.92 & 0.92 & 0.92 & 0.86 & 0.85 & 0.86 & 0.85 \\ \hline
\multicolumn{1}{|c}{CIFAR-10} & Clean & No & 0.74 & 0.74 & 0.69 & 0.74 & 0.93 & 0.92 & 0.93 & 0.93 & 0.87 & 0.86 & 0.87 & 0.86 \\ \hline
\multicolumn{1}{|c}{CIFAR-10} & Poisoned & Yes & 0.74 & 0.73 & 0.73 & 0.73 & 0.91 & 0.91 & 0.91 & 0.91 & 0.85 & 0.85 & 0.85 & 0.85 \\ \hline
\multicolumn{1}{|c}{CIFAR-10} & Clean & Yes & 0.74 & 0.74 & 0.74 & 0.74 & 0.93 & 0.93 & 0.93 & 0.93 & 0.86 & 0.86 & 0.86 & 0.86 \\ \hline
\end{tabular}}
\caption{Accuracy before and after clean retraining,  for Flowers and CIFAR-10.}
\label{tab:main_table_accuracy}
\end{center}
\end{table*}

\subsection{Effect of Retraining on Persistence}
We next look at the extent to which different architectures retain the backdoor even after retraining on clean data. We take each of the models described in the previous experiments and fine-tune (``retrain'') them on the smaller, untampered-with clean training set. The results are aggregated analogously and reported in the ``Retrained? Yes'' rows of Tables \ref{tab:main_table_advsucc} and \ref{tab:main_table_accuracy}.  %
These results show that clean retraining is an effective method for unlearning adversarial features. ResNet50, NasNet and NasNet mobile's adversarial test accuracy decrease significantly while model accuracy (on either clean or poisoned) is not affected. However, even after retraining NasNet still has almost 20\% adversarial success on square trigger pattern, far above ResNet50 and NasNet-Mobile. Therefore, the model developer's decision on architecture may have significant implications on performance as well as safety of the model.

\subsection{Effect of Regularization}
For all regularization experiments we use the simple black square trigger pattern and the Flowers dataset (the most effective pattern for the dataset). For each of five regularization strategies and for each of the five possible poison-classes in the Flowers dataset we train 10 ResNet50 models with different random samples of poisoned images, holding all hyperparameter choices constant. The initial weights of the models are pre-trained on the ImageNet image classification task provided from Pytorch model zoo \cite{model_zoo}. We use a learning rate of 0.00001, mini-batch size of 32, and Adam optimization to train all models.

\begin{figure}
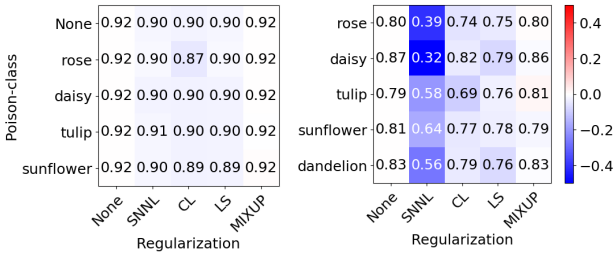

    \centering
    \begin{subfigure}[b]{0.25\textwidth}
        \includegraphics[width=0.87\textwidth]{pics/acc.png}
        \caption{Clean validation accuracy}
        \label{fig:valacc}
    \end{subfigure}%
    ~\hspace{-20pt}
    \begin{subfigure}[b]{0.25\textwidth}
        \includegraphics[width=1.0\textwidth]{pics/advacc.png}
        \caption{Adversarial success rate}
        \label{fig:advsuccess1}
    \end{subfigure}
    \caption{Average clean validation accuracy and adversarial success rate over 10 experimental runs, with many-to-one poison-class strategy. The color bar shows the difference relative to no regularization (column 1).}
    \label{fig:acctables}
\end{figure}

Figure \ref{fig:acctables} shows two tables of results: (a) accuracy on the clean validation set, which both the developer and adversary would like to maximize and (b) adversarial success rate, which the adversary would like to maximize but the developer would like to minimize. The columns of the tables correspond to
the regularization strategy employed and the rows correspond to the poison-class. The color of each cell indicates the difference regularization has relative to no regularization (column 1). Blue indicates that regularization decreases the value.
We see in Fig.~\ref{fig:valacc} a marginal drop in clean validation accuracy for all regularization strategies except for Manifold Mixup which does not affect performance on the validation set. The largest drop in validation accuracy comes from using the contrastive loss with Rose as the poison-class. Fig.~\ref{fig:advsuccess1} shows that SNNL, Contrastive, and Logit Squeezing regularization strategies all have the effect of lowering average adversarial success rates. However, SNNL has a more dramatic effect, dropping the overall average adversarial success rate across all poison-classes by 31\% absolute (from 82\% to 51\%). Note also that the poison-class has little effect on accuracy, but significantly affects adversarial success.

\begin{figure}
    \centering
    \quad %
    \begin{subfigure}[t]{0.25\textwidth}
        \includegraphics[width=0.95\textwidth]{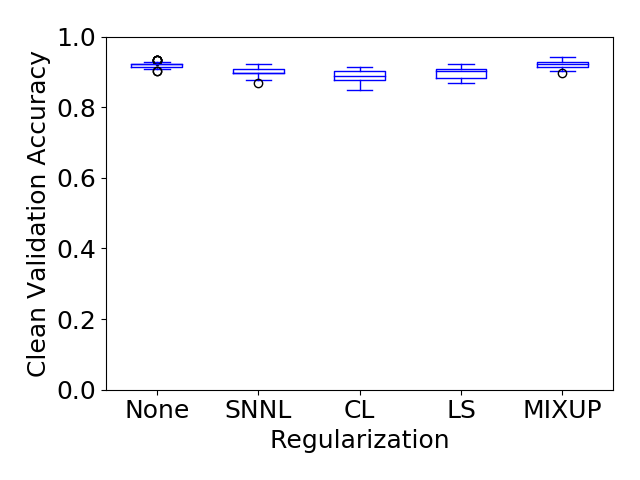}
        \caption{Clean validation accuracy}
        \label{fig:boxacc}
    \end{subfigure}%
    ~\hspace{-5pt}
    \begin{subfigure}[t]{0.25\textwidth}
        \includegraphics[width=0.95\textwidth]{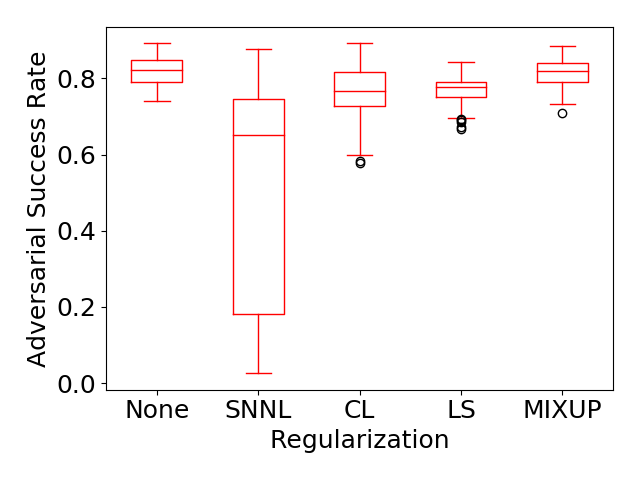}
        \caption{Adversarial success rate}
        \label{fig:advsuccess2}
    \end{subfigure}
    \caption{Validation accuracy and adversarial success rate as a function of regularization strategy.} \label{fig:box} 
\end{figure}

To get a sense of the consistency of these findings, Fig.~\ref{fig:box} shows the spread of validation accuracy and adversarial success rate across the 50 experimental runs for each regularization strategy, as a box-and-whiskers plot.
We see that all regularization strategies besides Manifold Mixup have a more dramatic affect on adversarial success rate than validation accuracy. The variance for adversarial success rate with SNNL loss is quite a bit larger compared to the other regularization methods. We conclude that regularization can be used to defend a model without significantly degrading the baseline performance on the validation set. 

\subsection{Effect of Trigger Pattern Transparency}
\begin{figure*}
\centering
\begin{tabular}{ccc}
\subcaptionbox{Low Variance,  Flowers}{\includegraphics[width = 2in]{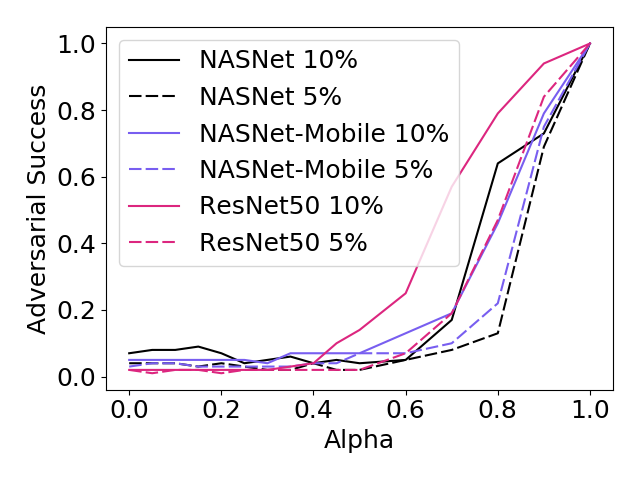}} &
\subcaptionbox{Sine, Flowers}{\includegraphics[width = 2in]{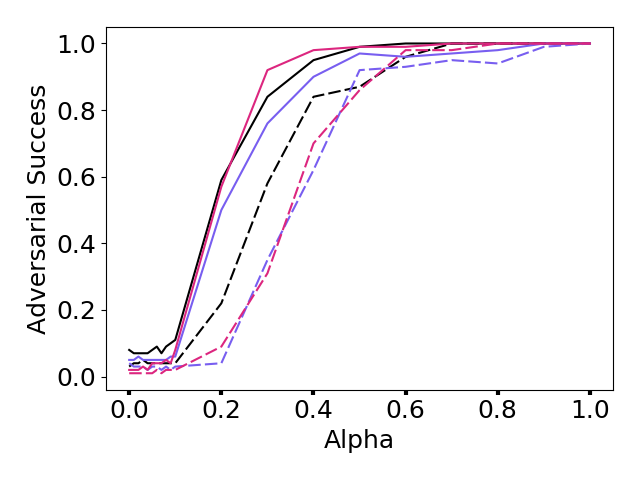}} &
\subcaptionbox{After clean retraining, Flowers}{\includegraphics[width = 2in]{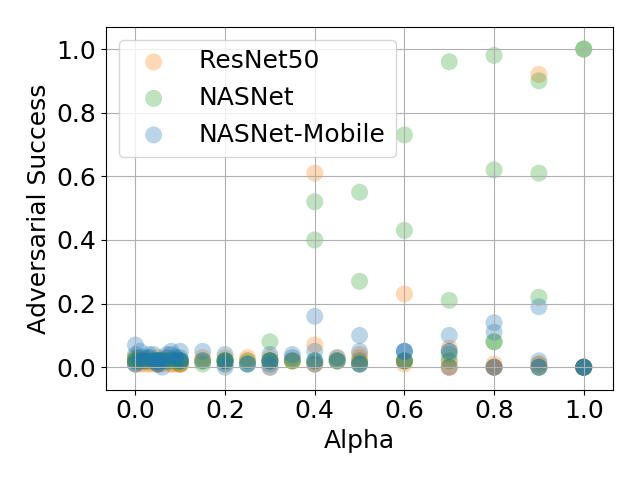}} \\
\subcaptionbox{Low Variance, CIFAR-10}{\includegraphics[width = 2in]{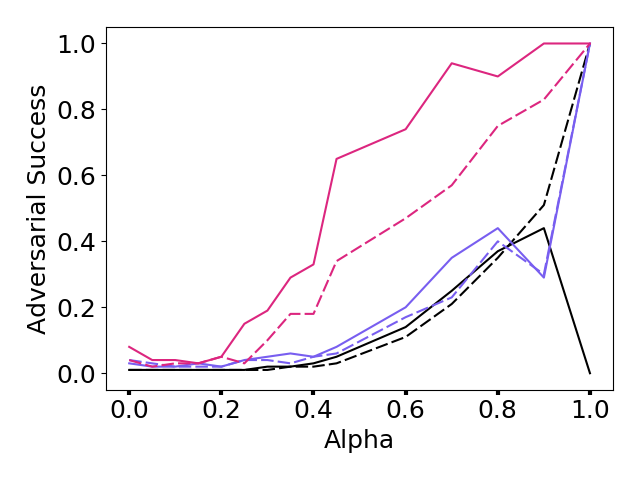}} &
\subcaptionbox{Sine, CIFAR-10}{\includegraphics[width = 2in]{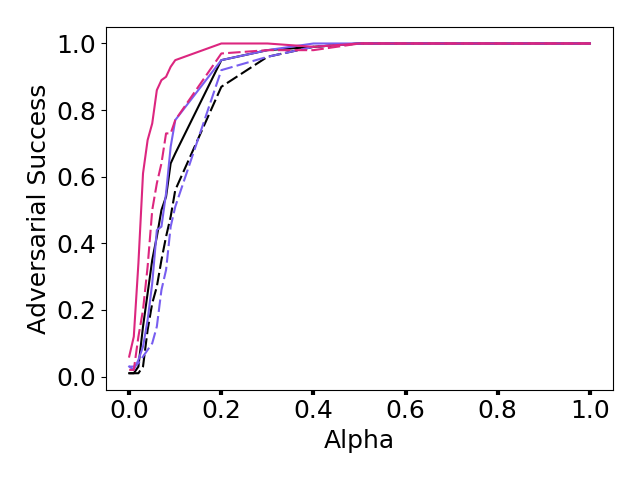}} &
\subcaptionbox{After clean retraining, CIFAR-10}{\includegraphics[width = 2in]{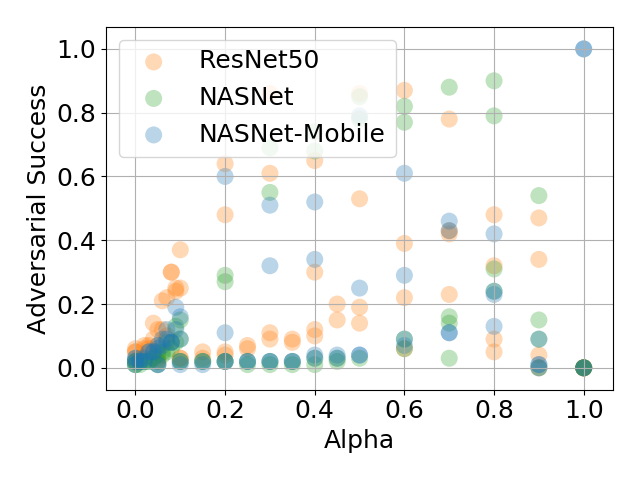}} 
\end{tabular}
\caption{Effect of $\alpha$.}
\label{fig:alpha}
\end{figure*}

Here we address effect of the trigger pattern transparency parameter, $\alpha$. Because square and random square use $\alpha=1$, we limit this analysis to the sine and low variance triggers.
We concentrate the range of tested $\alpha$ values on the lower range, since higher $\alpha$'s are less realistic. We also only target the most robust poison-classes, truck and rose, for CIFAR-10 and Flowers, respectively. These experiments compare  poison-rates of $\lambda=0.05$ and $\lambda=0.1$, with a total of 432 runs. Our results are shown in Fig.~\ref{fig:alpha}, the top row using Flowers and the bottom row CIFAR-10.
We find that higher $\alpha$ values can increase the trigger's effectiveness significantly, although the most effective performance comes when the trigger pattern is clearly perceptually detectable to humans. However, safety concerns remain because high $\alpha$ but low poison-rate attacks may be feasible in a big dataset where manual inspection of even a fraction of the samples is impractical. Figs.~\ref{fig:alpha}c and \ref{fig:alpha}f show performance after retraining with clean data, finding that retraining is not always effective against full image trigger patterns at sufficiently high $\alpha$.
A comprehensive defensive strategy should include a mechanism to detect "obvious" samples perturbed with high alpha triggers. We attribute the 0\% adversarial success at $\alpha=1.0$ in Fig.~\ref{fig:alpha}d to two factors: 1) at $\alpha=1$, all poisoned samples are identical and thus 0\% and 100\% are the only valid outcomes, and 2) noise in the training process at the particular early stopping point.

\subsection{Effect of Poison-rate}

\begin{figure}
\centering
\begin{tabular}{c}
\subcaptionbox{NasNet's accuracy on CIFAR-10 as a function of poison-rate, ranging over all trigger patterns. }{\includegraphics[width = 2.3in]{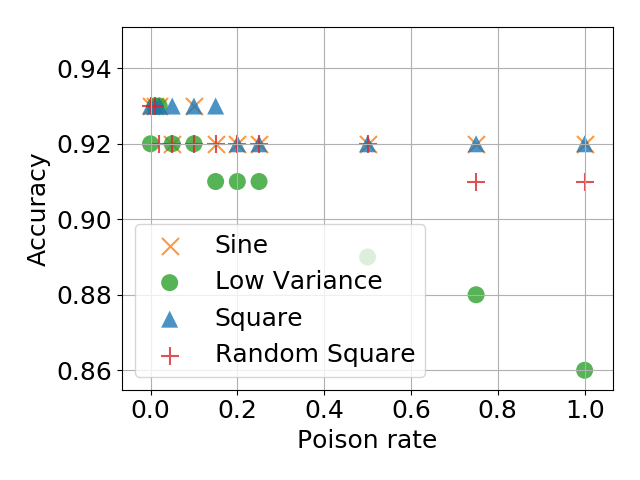}} \\ \subcaptionbox{NasNet's adversarial success on CIFAR-10 as a function of poison-rate.}{\includegraphics[width = 2.3in]{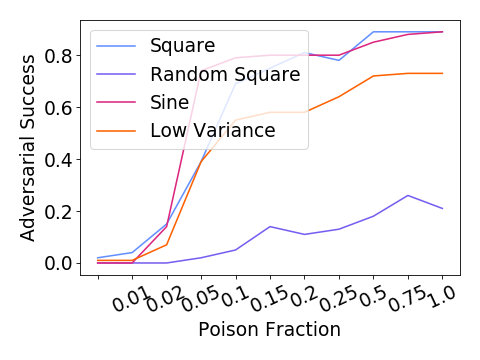}}
\end{tabular}
\caption{Effect of poison-rate.}
\label{fig:effectOfPoisonFraction}
\end{figure}

We next study the effect of poison-rate, $\lambda$.
We used CIFAR-10 as it has more samples per class than the Flowers dataset, providing us finer granularity for the poison-rate. 
Once again, we only target truck (the most robust poison-class on CIFAR-10) and focus primarily on small $\lambda$ values because they are more practical. Here we use only NasNet, since it has the highest clean and poisoned validation accuracy on CIFAR-10 using our standard hyper-parameters with a total of 44 runs.
Unsurprisingly, 
Fig.~\ref{fig:effectOfPoisonFraction}a shows that accuracy on poisoned validation steadily decreases as the poison-rate increases (as the poison-rate increases, the number of actual training samples in the target class decreases).
Fig.~\ref{fig:effectOfPoisonFraction}b plots the adversarial success rate as a function of poison-rate for different trigger patterns.  Sine requires the least poisoning, as it is extremely effective even with 2\% poisoning.  Random square requires the most poisoning, only finding middling success with impractically high poisoning rates.

\begin{table}
\centering
\resizebox{0.47\textwidth}{!}{%
\begin{tabular}{|cc|cc|cc|}
\hline
\multicolumn{2}{|c|}{\textbf{Accuracy}} & \multicolumn{2}{c|}{Poisoned} & \multicolumn{2}{c|}{Clean} \\ \hline
Model & Dataset & 1-to-1 & M-to-1 & 1-to-1 & M-to-1 \\ \hline
ResNet50 & Flowers & $0.87 \pm 0.01$ & 0.89 & $0.90 \pm 0.01$ & 0.88 \\ \hline
NasNet & Flowers & $0.86 \pm 0.02$ & 0.89 & $0.85 \pm 0.01$ & 0.86 \\ \hline
NasNet-M & Flowers & $0.78 \pm 0.02$ & 0.82 & $0.81 \pm 0.03$ & 0.84 \\ \hline
ResNet50 & CIFAR-10 & $0.71 \pm 0.03$ & 0.70 & $0.71 \pm 0.02$ & 0.69 \\ \hline
NasNet & CIFAR-10 & $0.92 \pm 0.01$ & 0.92 & $0.93 \pm 0.00$ & 0.93 \\ \hline
NasNet-M & CIFAR-10 & $0.85 \pm 0.01$ & 0.85 & $0.85 \pm 0.01$ & 0.86 \\ \hline
\end{tabular}}
\caption{Accuracy for one-to-one vs many-to-one.}
\label{tab:1to1VmanyVal}
\end{table}

\begin{table}
\centering
\resizebox{0.35\textwidth}{!}{%
\begin{tabular}{|ll|ll|}
\hline
\multicolumn{2}{|l|}{\textbf{Adversarial Success}} & \multicolumn{2}{l|}{Adversarial Test} \\ \hline
Model & Dataset & 1-to-1 & M-to-1 \\ \hline
ResNet50 & Flowers & $0.54 \pm 0.02$ & 0.72 \\ \hline
NasNet & Flowers & $0.37 \pm 0.02$ & 0.71 \\ \hline
NasNet-M & Flowers & $0.13 \pm 0.14$ & 0.35 \\ \hline
ResNet50 & CIFAR-10 & $0.58 \pm 0.18$ & 0.97 \\ \hline
NasNet & CIFAR-10 & $0.27 \pm 0.07$ & 0.73 \\ \hline
NasNet-M & CIFAR-10 & $0.40 \pm 0.11$ & 0.85 \\ \hline
\end{tabular}}
\caption{Adversarial success for one-to-one~vs~many-to-one.}
\label{tab:1VSmanyADV}
\end{table}

\subsection{One-to-one vs Many-to-one} \label{sec:1to1}
Lastly, we evaluate whether the one-to-one (``1-to-1'') or many-to-one (``M-to-1'') poisoning strategy is more effective. Table~\ref{tab:1to1VmanyVal} compares the accuracies of these two strategies for all models on both datasets. 
Square is used to poison Flowers while Sine is used to poison CIFAR-10 (the most effective patterns for them, respectively). 
Recall that one-to-one and many-to-one use the same number of poisoned samples for a given poison-rate;  the only difference is the source of the poisoned samples. 
The table reveals that these poisoning strategies do not have a significant impact on either the poisoned or clean validation set accuracies.
In contrast, Table~\ref{tab:1VSmanyADV} shows that many-to-one is significantly more effective than one-to-one in terms of adversarial success.
We hypothesize this is because the model incorporates the adversarial features better when the trigger pattern is spread across many classes, all pointing to the same target class.

\section{Conclusions and Future Work}
\label{sec:Conclusion}
This paper presents a systematic study of backdoor poisoning attacks on image classifiers. We evaluate the effect of design decisions within the model developer's control, including model architecture, regularization scheme, and any additional fine-tuning on a smaller, clean dataset, as well as those within the control of an adversary, including the trigger pattern and the rate and strength of the poisoning. We evaluate these on two datasets, Flowers and CIFAR-10, to assess the sensitivity to the particular training task. We report four key findings:
\begin{enumerate}
    \item \setlength{\parskip}{-2pt} Adversarial success rate varies widely depending on several factors, including model architecture, trigger pattern and regularization technique.
    \item While one would expect model performance and adversarial success to be negatively correlated, we find this rarely to be the case, suggesting poisoned models are not detectable through performance inspection alone. 
    \item Regularization typically reduces backdoor success rate, although it can have no effect or even slightly increase it, depending on the form of regularization. 
    \item Backdoors inserted through data poisoning can be rendered ineffective after just a few epochs of additional training on a small set of clean data without affecting the model's performance.
\end{enumerate}

We intend our current assessment to serve as a resource for safe and effective model development practices in face of adversity. 
However, adversarial machine learning is a rapidly evolving field of research. Backdoor data poisoning assessment can be characterized as the analysis of a two player zero-sum game with emerging innovative actions for the roles of adversary and developer, and so a complete analysis is beyond the scope of any single research study.

For future work, one could extend our assessment along three complementary dimensions. First, one could explore a greater range of values for studied factors (Fig.~\ref{fig:factors});
\eg, assessing with a larger dataset such as ImageNet.
Recent work also motivates additional regularization methods, such as Gaussian mixture loss \cite{yaseen2020preventing} and  $\ell_2$ regularization \cite{carnerero2020regularisation}, which can also partially mitigate data poisoning attacks.
Second, there are further factors of model developer decisions influencing model behavior which should be explored. To our knowledge, the choice of optimizer (\eg, SGD, Adam, AdamW \cite{loshchilov2017decoupled}) has not been evaluated in the context of backdoor data poisoning.
Lastly, one could extend our assessment of adversarial exploits. For instance, in this work we assess attacks which falsely label images, but {\em clean-label} backdoor attacks without label alteration have recently been demonstrated \cite{DBLP:journals/corr/abs-1902-11237, turner2018clean, saha2019hidden, zhu2019transferable}. 

\clearpage
\balance
\bibliographystyle{ieee_fullname}
\bibliography{new}

\end{document}